# IMPROVING TRANSLATION INVARIANCE IN CONVOLUTIONAL NEURAL NETWORKS WITH PERIPHERAL PREDICTION PADDING

*Kensuke Mukai and Takao Yamanaka*

Department of Information and Communication Sciences, Sophia University, Japan

## ABSTRACT

Zero padding is often used in convolutional neural networks to prevent the feature map size from decreasing with each layer. However, recent studies have shown that zero padding promotes encoding of absolute positional information, which may adversely affect the performance of some tasks. In this work, a novel padding method called Peripheral Prediction Padding (PP-Pad) method is proposed, which enables end-to-end training of padding values suitable for each task instead of zero padding. Moreover, novel metrics to quantitatively evaluate the translation invariance of the model are presented. By evaluating with these metrics, it was confirmed that the proposed method achieved higher accuracy and translation invariance than the previous methods in a semantic segmentation task.

## 1. INTRODUCTION

In recent years, some researches have focused on padding techniques, one of the most fundamental components in convolutional neural networks (CNNs). Although the padding changes only a few pixels at the edges of feature maps, this padding sometimes affects recognition accuracy in deep neural networks. For the padding, zero padding has been generally used due to low computational cost and simplicity. However, recent studies suggest that the use of zero padding promotes the encoding of positional information in CNNs [1, 2, 3, 4, 5] and may hinder the acquisition of translation invariance [6, 7, 8].

Encoding positional information in a network helps achieve high accuracy for the tasks where positional information is useful for identification, such as object recognition in an image [9]. On the other hand, in some tasks such as semantic segmentation, it has been confirmed that biased estimation based on absolute positional information tends to underestimate objects at the edge of an image [10, 11, 12], or that blind spots appear where objects cannot be recognized [13]. Such a phenomenon becomes a problem when learning with datasets such as satellite images where positional information (position in the cropped image coordinates) has no information about the segmentation label, or when using it for practical applications such as autonomous driving.

The previous methods proposed to deal with these problems aim to improve accuracy by padding image edges with more natural distributions [8, 14, 15] or by preventing specific patterns from appearing at image edges [9]. They are based on the idea that an unnatural distribution of image edges due to zero padding encourages CNN to encode positional information and adversely affects the performance of some tasks. However, in these methods, it is necessary to design an appropriate padding depending on the target task though the improvement of accuracy has been limited.

In this paper, a novel padding method was proposed, called Peripheral Prediction Padding (PP-Pad). PP-Pad estimates the optimal padding values from the values of several neighbouring pixels by end-to-end training of the model. Moreover, novel evaluation metrics were defined to evaluate the translation invariance of CNN models with each padding method. Using the metrics, the proposed padding methods were evaluated on estimation accuracy and translation invariance, compared with the previous padding methods. As a result of experiments in a semantic segmentation task, the proposed method achieved better recognition accuracy than that of previous methods with higher translation invariance.

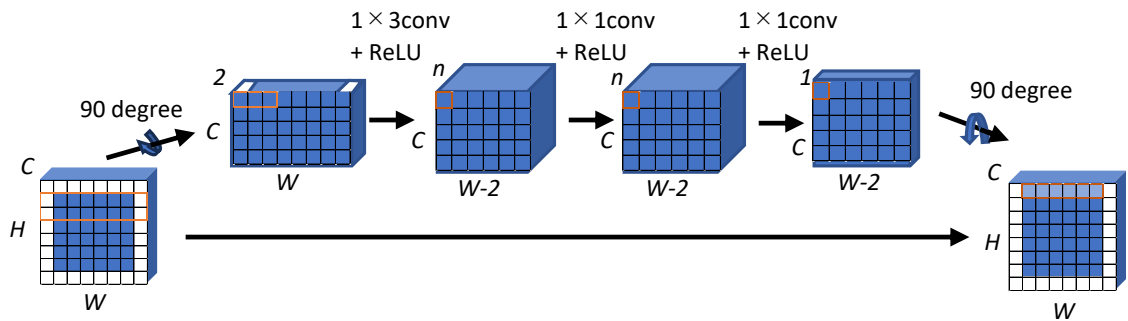

Fig. 1 Implementation of PP-Pad with convolutional layers ($h_p \times w_p = 2 \times 3$)

## 2. RELATED WORK

Some recent studies have conducted experiments on the effects of padding in convolutional neural networks [7, 14, 15, 16]. For the padding, zero padding has been commonly used, simply filling the edges of the image with 0. However, recent studies suggest that this zero padding facilitates the encoding of positional information in CNNs, and can reduce translation invariance, which is one of the important properties of CNNs [6]. The conventional padding methods other than the zero padding include replicate and reflect, which repeat values at the edge of the image. In addition, padding such as circular may be used in panorama images such as spherical images. Nguyen et al. have proposed a padding method that uses the mean and variance of surrounding patches and fills with values following a normal distribution [8]. Liu et al. have proposed a padding method that applies the image inpainting task using partial convolution [17, 18]. More recently, Huang et al. have proposed a model that treats padding as an image extrapolation problem and uses an image generation model to generate padding values [19]. In these methods, it has been considered that the padding values should have same distribution as the pixel values at the edges of an image. On the other hand, PP-Pad does not regularize the model to have the similar distribution in the padded pixel values to the edges of an image. It provides more flexible and more useful padding values for the models by estimating the optimal padded values for the task.

## 3. METHOD

### 3.1. Peripheral Prediction Padding (PP-Pad)

In the proposed method, the optimal padding value is learned in the end-to-end manner from the neighbouring $h_p \times w_p$ pixels using convolutional layers of $1 \times w_p$ kernel and 1×1 kernel, as shown in Fig. 1. When the input feature map $H \times W \times C$ with zero padding is given, the $h_p \times W$ region at the top edge except for the zero-padding region (orange square in the most left feature map in Fig. 1) is cropped to estimate optimal padding values at the top padding region (white pixels at the top edge). The cropped tensor $h_p \times W \times C$ is rotated 90 degrees to place the face of $C \times W$ in front as shown in the 2nd feature map in Fig. 1, and then is processed by a convolutional layer with a $1 \times w_p$, followed by an activation function ReLU (Rectified Linear Unit). After additional two 1×1 convolutional layers are applied with ReLUs, the feature map of $C \times (W - 2) \times 1$ is rotated back to the original direction, and then used as the padding values at the top edge of the original feature map except for the pixels at both ends, as shown in the rightest feature map in Fig. 1. Similar models are prepared separately for left, bottom, right sides for the feature map to estimate the optimal padding values. Note that the four corners of the feature map are padded with 0 in the proposed PP-Pad to avoid recursive predictions. Although the padding model could be designed with normal convolutional layers without the feature map rotation, the model size (the number of parameters) can be reduced by the proposed architecture. For example, if the padding model is designed with a normal convolution of a $h_p \times w_p = 2 \times 3$ kernel and two 1x1 convolutional layers, the number of padding model parameters is $(6Cn + n^2 + Cn)$, where $n$ is the number of intermediate channels. On the other hand, the proposed method is implemented with the $(6n + n^2 + n)$ parameters, so that the $7n(C - 1)$ parameters can be saved for each convolutional layer.

### 3.2. Evaluation Metric for Translation Invariance

In this work, the property of translation invariance for the proposed model was evaluated using a semantic segmentation task. Since there is no standard metric to measure the translation invariance, novel metrics were designed. As shown in Fig. 2, a patch is cropped from an image to estimate a class for each pixel of the patch using a semantic segmentation model with a padding method. Then, the patches are cropped in a sliding-window manner to estimate the class for each pixel. Since the cropped patches are overlapping, multiple class labels are obtained for each pixel in the original image. If the semantic segmentation model has the property of the translation invariance, these class labels are same for each pixel in the original image. Although the CNN-based model is thought to have translation invariance, it has been observed that the padding in the model provides positional information to better predict a class depending on the position in an image [12]. Therefore, the cropped patches sometimes provide different class labels for each pixel due to the padding. In this work, the translation invariance is measured by calculating the degree of coincidence over the predicted classes for each pixel obtained from the overlapping cropped patches. Specifically, this degree of coincidence is measured by entropy in Eq. 1.

$$e = -\sum_{k=0}^{K-1} p_k \log_2 p_k \quad (1)$$

$K$ is the number of classes, and $p_k$ is the probability of classifying a pixel into the $k$-th class, which is obtained from the histogram of the predicted classes for each pixel estimated from the multiple patches. The entropy $e$ takes a lower value for higher degree of coincidence. For example, $e$ is 0 if the predicted classes are same for all overlapping patches. Based on the entropy, two metrics for measuring the property of translation invariance are defined.

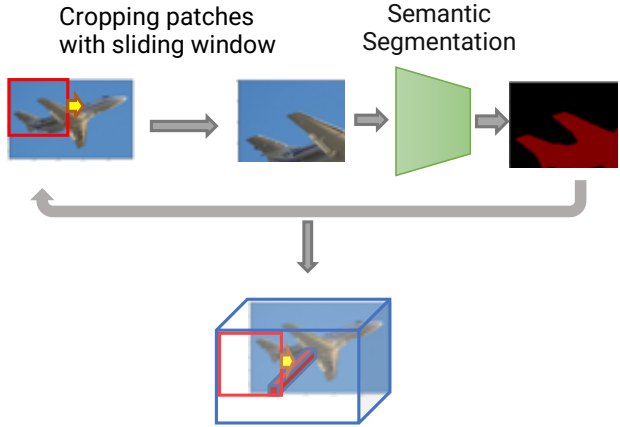

Fig. 2 Evaluation of translation invariance

$$meanE = \frac{1}{N}\sum_{n=0}^{N-1} e_n \quad (2)$$

$$disR = \frac{1}{N}\sum_{n=0}^{N-1} f_0(e_n) \quad (3)$$

$e_n$ represents the entropy for the *n*-th pixel, and $N$ is the total number of pixels in all images used for evaluation. Thus, *meanE* (mean entropy) represents the average entropy over all pixels. $f_\theta(x)$ is a function which takes 1 for $x > \theta$ and 0 for $x \leq \theta$. Thus, *disR* (disagreement rate) counts the number of pixels where the predicted classes are not same over overlapping patches. Therefore, both metrics in Eq. 2 and Eq. 3 take lower values for higher degree of coincidence, the property of translation invariance, and were used for the evaluation in the experiments.

## 4. EXPERIMENTAL SETUP

### 4.1. Dataset and Metrics

The dataset used in experiments is the PASCAL VOC 2012 dataset [20]. This is a dataset with 21 classes of people, animals, vehicles, and indoors, containing data for classification, detection, and segmentation. In the experiments, a segmentation dataset with 1464 training and 1449 validation data was used. During validation, patches were cropped from an image in a sliding-window manner. The number of patches input to the model was too large if all the validation images were used for the evaluation, which was computationally prohibitive. Therefore, 100 images out of the 1449 validation images were randomly selected and used as validation data.

The performance was evaluated using three metrics: *meanE*, *disR*, and mIoU (mean Intersection over Union) [21]. *meanE* and *disR* evaluate the translation invariance of the model with each padding, as previously described. mIoU was used for evaluating accuracy in the semantic segmentation task, by calculating it for all the patches obtained from a sliding window.

### 4.2. Training and Implementation Details

Unlike the previous methods [8, 17], it is not required for the proposed method PP-Pad to train a separate network to estimate padding values. In the experiments, Pyramid Scene Parsing Network (PSPNet) [21], an FCN model for semantic segmentation, was used as a base model. The initial values of the model were obtained from the weight parameters trained with the ADE20K dataset [22] and then trained with the PASCAL VOC2012 dataset using zero padding for 160 epochs. This aimed at reducing overall training time by the pre-training with the computationally efficient zero padding. After that, each padding method was applied to the model and trained it again for 160 epochs. Following [21], the "poly" learning rate policy [23] was used, where the current learning rate was equal to the base rate multiplied by $\left(1 - \frac{epoch}{maxepoch}\right)^{power}$. The base learning rate was set to 0.01 with *power* = 0.9. The momentum and the weight decay were set to 0.9 and 0.0001, respectively. Due to the limited GPU memory size, the batch size was set to 12 during training.

All images in the dataset were resized to 1050 pixels on the short side. During training, patches of $475 \times 475$ were randomly cropped and then were randomly flipped and rotated for data augmentation. When evaluating translation invariance and estimation accuracy, after resizing the image to 1050 pixels on the short side, patches of 475×475 pixels were cropped in a sliding-window manner with stride = 47 pixels. Since a same object can appear at the edges or the center of a patch cropped from an image due to the sliding window, positional information is not useful to predict the class for each pixel in this semantic segmentation task.

PP-Pad was implemented with $h_p$ = 2, 3, 5, 10, $w_p$ = 3, and *n* = 8. The size $h_p \times w_p$ represents the reference region to be used for calculating a padding value in PP-Pad. Note that the padding values were calculated in the channel-wise manner with a same convolutional filter among channels (Fig. 1).

## 5. EXPERIMENTAL RESULTS

The model was evaluated using three metrics: mIoU, *meanE*, and *disR*. While mIoU is a measure for recognition accuracy, *meanE* and *disR* are the two proposed measures for evaluating translation invariance. The results for the proposed method were compared with other padding methods, including the previous padding methods: CAP (Context-Aware Padding) [19] and Partial (Partial Padding) [17], as shown in Table 1. Both CAP and Partial aim to pad the edges of an image to have values from a natural distribution. As can be seen from the table, the proposed method achieved high recognition accuracy in mIoU with better translation invariance in *meanE* and *disR*. This would be because PP-Pad learned the model to produce the optimal padding values for the trained task. On the other hand, the padding methods such as reflect, replicate, and circular tended to have poor recognition accuracy and translation invariance. The recognition accuracies for previous methods, CAP and Partial, were comparable to the zero padding, but there was no improvement in the translation invariance. This indicates that CAP and Partial would differently predict the classes for pixels at the edges of a patch, from the prediction at the center of a patch, similar to the zero padding which is known that the positional information is encoded [12]. For the proposed method, PP-Pad with the reference region of

Table 1 Comparison in recognition accuracy (mIoU) and translation invariance (*disR*, *meanE*)

|  | Methods | mIoU ↑ | meanE ↓ | disR ↓ |
|---|---|---|---|---|
|  | Zero | 0.3323 | 1.8811 | 0.6482 |
|  | Reflect | 0.3040 | 32.8108 | 0.8374 |
|  | Replicate | 0.3059 | 1.8945 | 0.6545 |
|  | Circular | 0.3112 | 1.8413 | 0.6344 |
| Previous | CAP [19] | 0.3380 | 1.8764 | 0.6429 |
|  | Partial [17] | 0.3324 | 2.0035 | 0.6846 |
| Proposed | PP-Pad (2×3) | 0.3352 | **1.7995** | **0.6102** |
|  | PP-Pad (3×3) | 0.3472 | 1.8787 | 0.6465 |
|  | PP-Pad (5×3) | 0.3380 | 1.8568 | 0.6321 |
|  | PP-Pad (10×3) | **0.3486** | 1.8334 | 0.6322 |

Table 2 Comparison of computational cost during training and inference

|  | Methods | Training on GPU [s/10iter] | Inference on CPU [s/image] |
|---|---|---|---|
|  | Zero | 2.45 | 0.606 |
| Previous | CAP [19] * | 2.98 | 0.710 |
|  | Partial [17] | 2.98 | 0.646 |
| Proposed | PP-Pad (2×3) | 7.26 | 0.755 |
|  | PP-Pad (10×3) | 7.35 | 0.802 |

* CAP [19] requires additional pre-training.

2×3 showed the best translation invariance with comparable recognition accuracy with the zero padding and previous methods. PP-Pad with the 10×3 reference region achieved the best recognition accuracy with slightly lower performance on the translation invariance than PP-Pad (2×3). The reason of the high recognition accuracy in PP-Pad (10×3) would be that more accurate padding values can be predicted from a larger area of an image.

Since the zero padding fills the edges of an image with zero values, the convolutional filters can recognize the edges by learning "boundary detector" which detects zero patterns at the edges. In addition, the values at the edges of the feature map are affected by zero values in every convolutional process. Therefore, the region where zero padding affects propagates to the center of the patch as convolutional layers become deeper. Thus, the zero padding would encode positional information to promote the recognition accuracy, which would lead to the decrease of the performance on translation invariance.

An example of the results on this semantic segmentation task is shown in Fig. 3. Although the original image includes almost whole body of a dog, a cropped patch indicated by the red frame in the ground-truth (GT) image, only includes a part of the dog. Since the shape of the dog in the cropped patch is similar to a bird, some pixels in the patches were recognized to the bird class (yellow) in CAP, Partial, and Zero, while PP-Pad accurately classified all the pixels to the dog class (purple) in this example.

Since it would be important to compare the computational cost during training and inference, the comparison of the computational costs is shown in Table 2 for the zero padding, the previous methods, and the proposed methods. From the table, the overhead in the proposed method for inference was relatively small, although the training required almost three-times training time compared with the other methods. It is noted that CAP requires additional pre-training of the padding model to obtain padding values. Since PP-Pad learns the padding model to produce the optimal padding values in the end-to-end manner, it took more time to train the model. However, PP-Pad can use a pre-training procedure with zero padding for computational efficiency. Therefore, the required time to obtain final model would be reduced for PP-Pad.

## 6. CONCLUSIONS

In this paper, a novel padding method, PP-Pad, was proposed to improve translation invariance in convolutional neural networks. In PP-Pad, the network can be learned in the end-to-end training to obtain the padding model to produce the optimal padding values. In order to evaluate the model, two metrics were defined as measures for translation invariance based on a semantic segmentation task. From the experiments, it can be confirmed that PP-Pad achieved high recognition accuracy with better translation invariance in the semantic segmentation task.

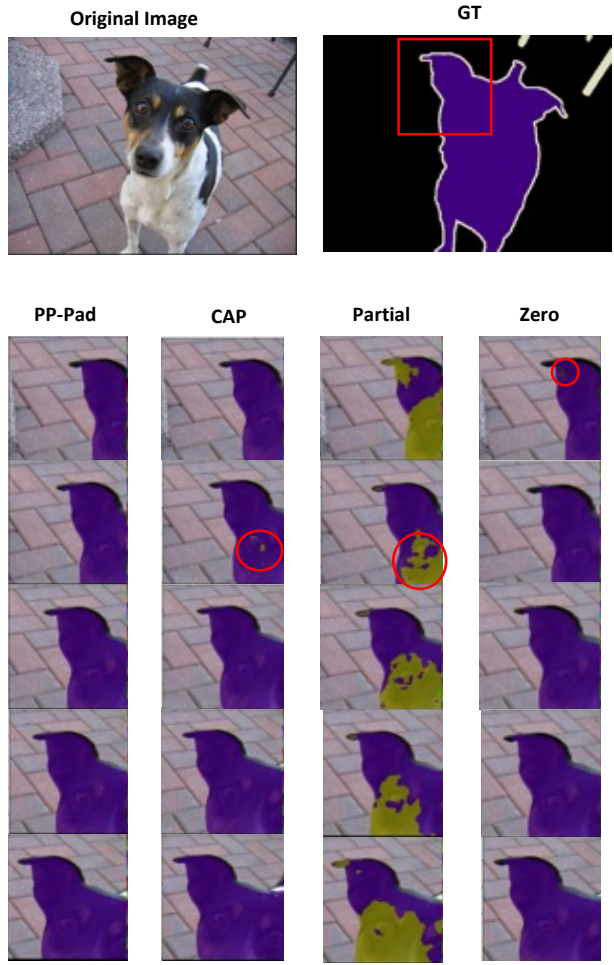

Fig. 3 Example of prediction results in semantic segmentation for sliding-window patches